%% file: main-archiv.tex
\documentclass[letterpaper,10pt,conference]{ieeeconf} 
\IEEEoverridecommandlockouts  
\usepackage{graphicx}
\usepackage{url}
\usepackage{xcolor}
\usepackage{amssymb}

\newcommand{\chk}{\checkmark}

\begin{document}

\title{\LARGE \bf Deep Smartphone Sensors-WiFi Fusion for\\
Indoor Positioning and Tracking}


\author{Leonid Antsfeld \hspace{12mm} Boris Chidlovskii \hspace{10mm} \quad Emilio Sansano-Sansano\\
Naver Labs Europe \hspace{9mm} Naver Labs Europe \hspace{7mm} Universitat Jaume I de Castello\\
\hspace{-15mm} France \hspace{30mm} France \hspace{35mm} Spain}

\maketitle

\begin{abstract}
We address the indoor localization problem, where the goal is to predict
user's trajectory from the data collected by their smartphone, using inertial sensors such as accelerometer, gyroscope and magnetometer, as well as other environment and network sensors such as barometer and WiFi. 
%
Our system implements a deep learning based pedestrian dead reckoning (deep PDR) model that provides a high-rate estimation of the relative position of the user. Using Kalman Filter, we correct the PDR's drift using WiFi that provides a prediction of the user's absolute position each time a WiFi scan is received. 
Finally, we adjust Kalman Filter results with a map-free projection method that takes into account the physical constraints of the environment (corridors, doors, etc.) and projects the prediction on the possible walkable paths.
	
We test our pipeline on IPIN'19 Indoor Localization challenge dataset and demonstrate that it improves the winner's results by 20\% using the challenge evaluation protocol. 
\end{abstract}


\input{introduction.tex}
\input{soa.tex}
\input{architecture.tex}
\input{evaluation.tex}

\section{Conclusion}
\label{sec:conclusion}
We propose a novel architecture for user's indoor localization based on data collected by a smartphone. We build a reliable prediction of the user's trajectory using inertial sensors such as accelerometer, gyroscope and magnetometer, as well as barometer and WiFi scanner. Our main innovation is a deep learning based pedestrian dead reckoning (PDR) model that provides a high-rate estimation of user's local displacement. We describe the full system and its components, including the landmark detection, relative and absolute position estimation from sensor data, prediction fusion map-free projection. We show how to shape sensor data to train CNNs/RNNs architecture. We evaluate our system on the IPIN'19 indoor localization challenge dataset and obtain the localization error which outperforms the challenge winner's performance by 20\%.

\bibliographystyle{plain}
\bibliography{references} 


\end{document}

%% file: introduction.tex
\section{Introduction}
\label{sec:introduction}


Ubiquitous location-based services have recently attracted a great deal of attention. They require a reliable positioning and tracking technology for mobile devices that works outdoors as well as indoors~\cite{zafari2017a}.
While navigation satellite systems such as GPS already provide reliable positioning outdoors, a corresponding solution is yet to be found for indoor environment 
where GPS signal cannot penetrate and provide sufficient accuracy performance.

Indoor location-based services~\cite{gressmann10towards} 
bring important social and commercial values, by enabling many applications including human localization and tracking, personalized advertisement, living assistance, etc. 
The ubiquity 
of smart-phones 
and the availability of different wireless infrastructure, such as WiFi and Bluetooth, 
make them an attractive platform for such positioning systems. 

Numerous techniques for smartphone-based indoor positioning  have been developed, yet there is not a single solution that can guarantee a reliable and universal service~\cite{wang-2019-survey} on its own. 
Most techniques exhibit their strengths and weaknesses under different conditions. In combination, they can complement each other and improve not only accuracy but also reliability of service. 

Nowadays a typical smartphone contains a dozen of different sensors, and their number keeps growing.
There are several types of sensors in a smartphone.
{\it Network sensors}, such as WiFi and Bluetooth, may be leveraged to estimate an absolute position of a user. WiFi positioning using received signal strength (RSS) fingerprinting~\cite{Ma2019wifi} have been considered as the most popular indoor positioning solutions. 
RSS values from several access points (APs) can be easily gathered by common smartphones under existing WiFi infrastructure. However, severe RSS fluctuations always render inaccurate positioning results. {\it Motion and Position sensors}, (a.k.a. {\it Inertial Measurement Unit, IMU}) such as accelerometer, gyroscope and magnetometer can help estimate user displacement relative to a known starting point.
This approach known as {\it pedestrian dead reckoning} (PDR) system~\cite{harle2013}, 
where PDR determines the user's location by adding the currently estimated displacement to previously estimated location. The displacement is estimated by combining step detection, step length estimation with user heading estimation from accelerometer, gyroscope and magnetometer data streams. PDR can achieve accurate positioning over short distances but is a subject of drift over long distance.
{\it Environment sensor} such as barometer, for example, may be useful in determining the floor inside a building. 
Both WiFi and PDR have serious limitations though
(high variation of WiFi signals and the drift of PDR) ~\cite{Ma2019wifi,wang-2019-survey},
so an auxiliary tool for indoor localization has been proposed. Namely, {\it landmarks} can be easily identified based on specific sensor patterns in the environment~\cite{chen-2015-fusion}, and then be exploited to correct WiFi and PDR predictions.  

In particular, human motion recognition~\cite{lima19human,wang20deep} from smartphone sensors may be used to improve indoor positioning. User motion states, like staying still, walking or taking stairs can be treated as indoor landmarks~\cite{abdelnasser16} to reset location estimation by PDR.
Therefore, recent works tend to fuse WiFi positioning, PDR and landmarks to enhance the indoor positioning accuracy~\cite{deng-2016-continuous}.

\subsection{Our proposal} 
In this paper, we propose a new sensor fusion framework for accurate indoor positioning and tracking using the smartphone inertial sensors, WiFi measurements and landmarks.
Our framework integrates new components which distinguish it from the state of art approaches and they are as follows:
\begin{enumerate}
\item{\it Deep PDR.}
Inspired by using deep learning for user's activity detection~\cite{deng-2016-continuous,lima19human}, we apply deep learning approach to PDR. 
We pre-process and reshape sensor data streams and 
use convolutional (CNN) and recurrent (RNN) networks to extract underlying hidden correlations between different sensors and modalities to learn a model of user local displacement. 
%





This allows to cope with sensor noise and replace the manual feature extraction which is a frequent subject to data noise and sophisticated thresholding, including tuning to different pedestrian profiles, depending on gender, age, height etc.~\cite{wang_pedestrian_2019}. 



While this approach gives a better relative displacement of the user, since the sensors measurements are always noisy,
we use WiFi based predictions and observed landmarks in order to obtain an absolute position of the user. 


\item {\it Landmarks and pseudo labels.}
CNN/RNN models require a large annotated dataset for training, while genuine ground truth annotations are sparse and available for a limited number of landmarks. On the other hand, raw sensor data are massively generated at a high rate. So we annotate sensor data with pseudo labels and generate a large annotated set for training CNN/RNNs. It is based on simpler tasks of user walking and landmark detection and a interpolation of user's behaviour between the landmarks.

\item {\it Semi-supervised VAE for WiFi.} 
A radiomap/fingerprinting is constructed from the WiFi data provided in training and validation data. Recorded data provides a WiFi scan reading every 4 seconds approximately, however without an exact position where this scan was taken. Using the provided inertial sensor data, we can infer the approximate position where the WiFi fingerprint was taken and build a radiomap with this information.

\end{enumerate}

The preliminary version of our framework participated in the off-site smartphone based positioning track of the competition organized at IPIN 2019 conference~\cite{ipin_competition_2019} and was ranked  2\textsuperscript{nd}. The full framework presented in this paper improves our own results by 25\%. Moreover, as evaluations show, it reduces the localization error obtained by the IPIN'19 challenge winner by 20\%.

The rest of the paper is organized as follows. Section 2 reviews the related work in WiFi based positioning, PDR based positioning and deep learning from sensor data. Section 3 presents the full architecture for indoor positioning and tracking. It then describes in detail the main components, paying a particular attention to deep PDR modeling, landmark recognition and pseudo labels for training CNN/RNN. Sections 4 presents the evaluation setting of IPIN'19 indoor localization challenge, and reports evaluation results and ablation studies. Finally, Section 5 concludes the paper.

%% file: soa.tex
\section{Related Work}
\label{sec:soa}

Several recent surveys give an exhaustive picture of different aspects of research in the mobile and wireless networking domains, including indoor positioning for smartphones
~\cite{davidson-2017-survey,harle2013,Ma2019wifi,zhang-2019-survey}. In this section, we briefly present works relevant to our architecture for indoor positioning and tracking, in particular, Wifi and PDR-based positioning, activity recognition and free-map matching.

\subsection{WiFi based positioning}

The most popular technique for smartphone-based indoor localization today is WiFi-fingerprinting 
\cite{davidson-2017-survey,gressmann10towards,Ma2019wifi,khalajmehrabadi-2017-modern}. 
A location is represented by a WiFi-fingerprint which lists visible access points and their respective received signal strength (RSS). Positioning is performed by matching the WiFi-fingerprint that is measured on the mobile device to a database of reference fingerprints collected beforehand during a calibration phase. The location associated with the closest match is returned as position estimate.

Over the past decade, most research effort focused either on improving the matching of measurements to reference data or on generalizing training data into signal strength models. 
Beyond a simple nearest neighbor matching~\cite{gressmann10towards}, 
Ferris et al.\cite{ferris07wifi} proposed to model signal strength across an entire building using Gaussian processes which allows to extrapolate to areas with no reference data. In contrast to that, 
~\cite{hilsenbeck_graph-based_2014} considered a continuous building-wide WiFi model unnecessary 
and propose a sparser representation by mapping fingerprints to a graph-based reference database.

To be accurate, the fingerprints should be densely recorded and annotated with exact coordinates.
Classical methods suffer from hand-crafted algorithms, subject of heavy complex calibration and parameter tuning.

The main challenge is a gap between a massive generation of non-annotated sensor data and their modest annotation allowing to deploy only simple machine learning algorithms~\cite{wang_pedestrian_2019}. To enable deployment of modern deep learning techniques, the problem of sizable annotations is usually addressed by crowd-sourcing, pseudo-labeling or semi-supervised learning able to combine unlabeled and labeled sensor data. Some efforts have been proposed for WiFi-based localization~\cite{chidlovskii19,mohammadi_semi-supervised_2018}.


In~\cite{yuan2014} Y. Yuan et al. introduced an efficient fingerprint training method, using semi-supervised learning, reporting 80\% time cost reduction while guaranteeing the localization accuracy. Another method was recently proposed in~\cite{Bi2019} where a faster radio-map construction is achieved by allowing larger distances between successive fingerprints and by using adaptive path loss model interpolation to estimate locations of fingerprints.
 

A semi-supervised method for localization of a moving smart-phone robot was proposed in~\cite{yoo17}. First they obtain pseudo labels for the unlabeled data using Laplacian Embedded Regression Least Square. During the learning phase, two decoupled balancing parameters are individually weighted to labeled and pseudo-labeled data. 
%
%
Semi-supervised learning with generative models based on Variational Auto-Encoder (VAE) has been applied to WiFi based localization in~\cite{chidlovskii19}; we deploy it in our positioning and tracking architecture.

\subsection{Pedestrian dead reckoning}
\label{ssec:pdr}

PDR-based localization technique utilizes 
the inertial sensors available on modern smartphones, 
in particular, accelerometer, gyroscope and magnetometer~\cite{harle2013}. 
%
As all inertial methods, it can give an accurate position only in a short period of time, but requires regular corrections of user's position to avoid the error accumulation. PDR is often composed of {\it step detection}, {\it step length estimation} and {\it heading determination}. The user's position is estimated recursively by accumulating vectors that represent the movement of the user at each detected step.

All PDR components are a subject of heavy parameter tuning~\cite{lima19human}.
Step length depends on user's characteristics such as height or age, and even for the same user, it may vary according to the activity the user is performing, i.e., walking slowly vs. walking fast.
Step detection algorithms, such as peak detection, flat-zone detection and zero-crossing detection, are not free of heavy parameter tuning either~\cite{shin-2007-adaptive}. Accuracy of these techniques depends on thresholds, that may be conditioned by the user's characteristics but also by the quality and particularities of the inertial sensors, being appropriately set
\cite{lee_2013_real}.

With respect to heading estimation, usually the heading angle offset, which is the angle between the direction of smartphone and the direction of the user, will not remain constant during the navigation. The assumption that the angle remains constant can be satisfied when pedestrians hold smartphones on the front of the body, but if the phone pose is arbitrary, the heading offset cannot be guaranteed to be constant.


Displacement and direction of motion are then estimated for individual steps. To this end, recent research relies on machine learning techniques and activity recognition~\cite{lima19human} has been extended from distinguishing not only between the user moving and standing still, but to further include estimating the walking speed, climbing on stairs, taking an elevator, etc.~\cite{wang_pedestrian_2019,zhou-2019-smartphone}. 
\subsection{Deep learning from sensor data}

A new generation of systems for indoor localization confirms a transition from traditional approaches of signal processing to machine learning solutions including deep learning~\cite{wang_pedestrian_2019,Ma2019wifi}. 

Most mobile devices can only produce unlabeled position data, therefore unsupervised and semi-supervised learning become essential. Mohammadi et al.
\cite{mohammadi_semi-supervised_2018} address this problem by leveraging deep reinforcement learning and variational auto-encoders (VAE). In particular, their framework envisions a virtual agent in indoor environments, which can constantly receive state information during training, including signal strength indicators, current agent location, and the real (labeled data) and inferred (via a VAE) distance to the target. 


Deep learning for recognition of human activities has been approached by using both ambient sensing methods and wearable sensing methods~\cite{zhou-2019-smartphone}. 

Activity recognition using sensor data is a multivariate time-series classification problem, which extracts discriminative features from sensor data to recognize activities by a classifier~\cite{lima19human}. As time-series data have a strong one-dimensional structure, in which the variables temporally nearby are highly correlated~\cite{wang-2019-survey,zhou-2019-smartphone}. Traditional methods 
rely on extracting complex hand-crafted features which require laborious human intervention and leads to the incapability of pedestrian activities identification.


In~\cite{zhou-2019-smartphone}, a deep learning-based method for indoor activity recognition by using the combination of data from multiple smartphone built-in sensors. A new convolutional neural network (CNN) has been designed for the one-dimensional sensor data to learn the proper features automatically. 


\subsection{Map-aided navigation}

The idea of using indoor space geometry for reduction of position and heading errors in autonomous positioning systems has been extensively exploited in the last several years. In the case of indoor navigation, building floor plans represent constraints that restrict movements, as people cannot walk through walls and floor changes can occur only via staircases or elevators. The goal of map-aided navigation is to exploit prior information contained in maps to improve positioning accuracy~\cite{davidson-2017-survey,perttula14}.
There are currently three approaches to map aided navigation indoors~\cite{tran17}, all of which can be implemented on smartphones probabilistic map matching based on particle filtering using wall constraints, topological map matching based on link-node representation of a building plan, and reduction of heading error by comparison with building cardinal heading. The purpose of these algorithms is to improve positioning and heading by adjusting the estimated path to the building plan~\cite{nguyen-huu17}.



%% file: architecture.tex
\begin{figure*}[ht]
\includegraphics[width=2\columnwidth,scale=1]{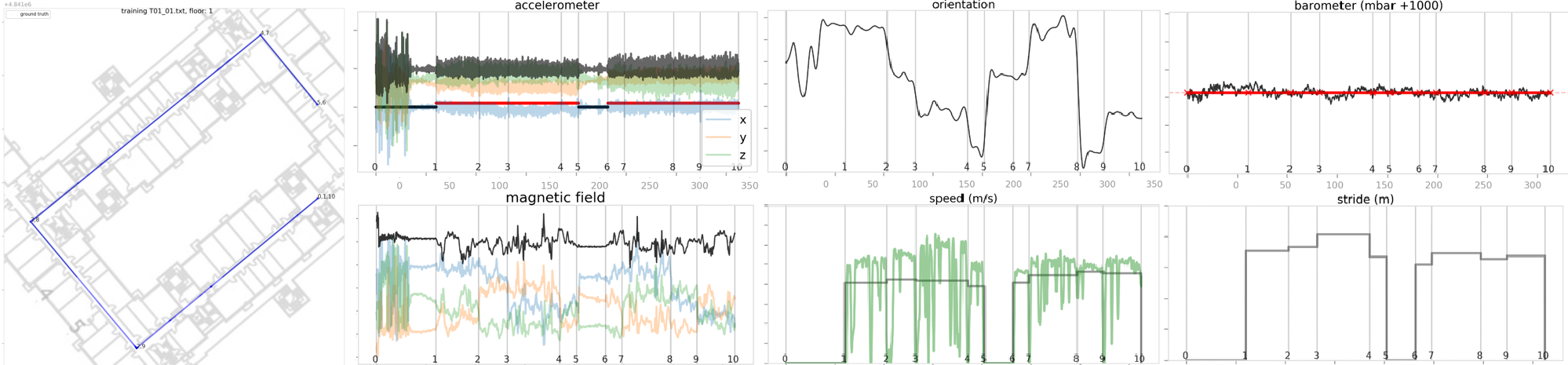}
\caption{Sensor data for landmarks detection.}
\label{fig:landmarks}
\end{figure*}

\section{System design} 
\label{sec:architecture}


We illustrate our approach using an example of user's route (see  Figure~\ref{fig:landmarks}) from IPIN'19 localization challenge dataset and associated data from accelerometer, gyroscope, magnetometer and barometer sensors, as well as speed and stride estimations. The route spans 10 points; it starts by switching the smart phone on at point 0 and letting the calibration terminate. The user then walks through points 1, 2, 3, 4 to point 5. Once at point 5, she returns (creating point 6) and walks through points 7, 8, 9 to get back to the starting point. The figure plots the sensor data streams along the timeline, through point 0 to 10.

Sensor data and landmarks 
are used to generate the pseudo labels for deep PDR learning.
The main elements of the sensor data annotation are the following:
\begin{itemize}
\item {\it Walking vs standing still}. 
A small amount of sensor data is sufficient to train an accurate classifier to distinguish between these two activities~\cite{wang-2019-survey}. In Figure~\ref{fig:landmarks}, accelerometer data between points 0 and 1 (after the calibration phase) and points 5 and 6 clearly suggests user's standing still. 
\item {\it Landmarks}. 
Points 1 to 5 and 7 to 10 of the route are landmarks, they refer to direction changes. Crucial for training indoor localization systems, they are commonly annotated with the ground truth positions. Figure~\ref{fig:landmarks} suggests that orientation changes\footnote{Orientation vectors can be estimated from accelerometer, gyroscope and magnetometer data either separately or via a smartphone application.} are highly correlated with landmarks. By coupling orientation data with other sensor data and landmark ground truth, a simple Random Forest can be trained to get the accurate landmark predictor~\cite{deng-2016-continuous}. 
\item{\it Pressure}. 
The user's route in Figure~\ref{fig:landmarks} stays on the same floor. In general, barometer sensor data allow to easily recognize the floor change~\cite{zhou-2019-smartphone}.
\item {\it Speed and Stride Estimations.} 
Obtained from accelerometer and gyroscope data, they are important to generate pseudo labels and annotations, and their values are inferred from PDR and averaged over each route's segment.
 \end{itemize}

Our assumption of steady user's walk between two landmarks is inspired by indoor localization datasets created for IPIN'18 and IPIN'19 challenges~\cite{ipin_competition_2018,ipin_competition_2019}. In a more general case with multiple open spaces and erratic user's walking, it can lead to over- or under-segmentation of a trajectory and a high noise in generated pseudo labels. 

\subsection{Deep Learning from sensor data}
\label{ssec:deep}

CNNs are state-of-the-art models in image recognition tasks, where the nearby pixels typically have strong relationships with each other thus forming visual patterns. 
While CNNs are used for computer vision tasks, we believe their convolutional layers are able to capture relationships in motion signals and identify correlations between sensors once the input is shaped as an image. In multi-modal approaches, where many sensors are used to capture a movement, grasping correlations among sensors may help to better interpret data. Thus CNN can exploit the local dependency characteristics inherent in time-series sensor data and the translation-invariant nature of movement.

To enable convolution on smartphone sensor data we frame it as an image. We first down-sample all raw sensor data to 50Hz, a frequency sufficient to characterize any user's displacement~\cite{zhou-2019-smartphone}. 
Then we implement two modes of converting sensor data, using raw data or recurrence plots.

In the {\it raw data} mode, we concatenate all sensor data in one $d$-dimensional stream and run a sliding window over the stream. 
The window width determines the width of each data point as CNN input, and represents the time interval considered. If the window width is set to one second and the data is sampled at 50Hz, each data point will be 50 columns wide. It is a rule of thumb in the community that a one second interval is adequate to characterize human activities and, therefore, it should be sufficient to learn a meaningful user's movement model.

\begin{figure}[ht]
	\includegraphics[width=\columnwidth, scale=1] {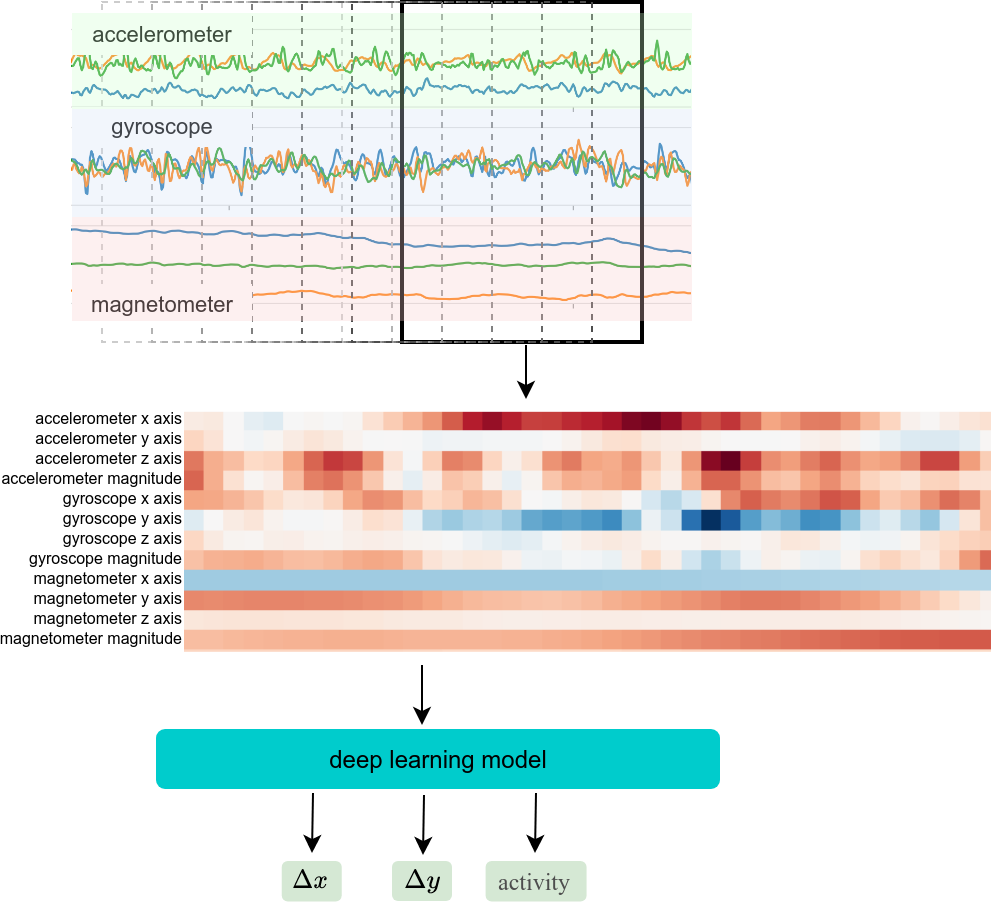}
	\caption{Framing raw sensor data as $12 \time 50$ input to CNN trained to predict the local displacement ($\Delta x, \Delta y$) and user activity.}
	\label{fig:sensor_features}
\end{figure}

The window height depends on the number of sensors that are being taken into account. For accelerometer, gyroscope and magnetometer sensors we generate four rows, one for each axis $x,y,z$ and the one for the magnitude, calculated from the three axial values. 
Figure~\ref{fig:sensor_features} illustrates this process. For each window considered, a total of 12 features are extracted for each time stamp and framed as an image.

We build the local displacement model as having one regression branch and one classification branch (see Figure~\ref{fig:sensor_features}). 
The regression branch is aimed at predicting user local displacement $(\Delta x,\Delta y)$, where the classification branch predicts the user activity. The activity classification is trained with the standard cross entropy loss; the regression branch is trained by minimizing the $L_2$ loss over a set of $N$ 2D points, defined as follows:
\begin{equation}
Loss_{regr}=\frac{1}{N}\sum_{n=1}^{N}||(\Delta \hat{x}_n,\Delta \hat{y}_n)-(\Delta x_n,\Delta y_n)||,
\label{eq:mae}
\end{equation}
where $(\Delta x,\Delta y)$ is a ground truth and $(\Delta \hat{x}, \Delta \hat{y})$ is a prediction. 

We train the deep PRD model using the \textit{Adam} \cite{kingma2014adam} optimizer with a learning rate $10^{-3}$ and a weight decay value $10^{-5}$. First, the raw sensor data is extracted from a set of annotated logfiles containing recorded values from all sensors and all tracks. A subset of these tracks is reserved for validation. The total training and validation losses are calculated as a weighted sum of $Loss_{regr}$ in (\ref{eq:mae}) and the cross entropy $Loss_{ce}$ for the activity the user is performing (standing still vs. walking):
\begin{equation}
 Loss_{total} = Loss_{regr} + \alpha Loss_{ce}
\label{eq:total} 
\end{equation}
where $\alpha$ is a trade-off between the two terms. In our experiments, we set $\alpha$ to 1. The training process stops when the validation loss has not improved for 50 epochs.

\subsection{Recurrence Plots}
\label{ssec:rp}
Recurrence-based analysis~\cite{goswami2019} utilizes a fundamental characteristic that any system eventually returns close to its earlier states as time passes. In the case of real-world time series, systems often repeat earlier behavior, even though they might at times be interrupted by regime shifts and dynamical transitions. Recurrence plots encode the pairwise recurrences of time series values and thus create a visual representation of system dynamics, solely from the measured time series. 

Consider $\{v_t\}^N_{t=1}$, a $d$-dimensional time series of length $N$. The system is said to recur when a state vector $v_i$ at time $t = i$ is close to a different state vector $v_j$ at time $t = j$, i.e., $v_i \approx v_j$. Here, the notion of $v_i$ being close to $v_j$ depends on (i) the choice of a norm such as the Euclidean norm or the maximum norm, and (ii) the choice of a distance threshold, which helps unambiguously define all states farther apart as 'not close', and vice versa. We can thus encode all possible pairs of recurrences in the recurrence matrix $R$, where
\begin{equation}
    R_{ij} = \Theta (\varepsilon - || v_i - v_j ||),
\label{eq:rp}
\end{equation}
$||\cdot||$ is a norm, $\varepsilon$ is a chosen distance threshold and $\Theta$ is a normalization function.

Working with the entire time series being not practical, we consider a window width $n \ll N$. The resulting matrix $R$ of size $n \times n$ is a matrix comprising solely values between 1 and 0 where the values close to 1 denote pairs of points where the sensor data recur, while values close to 0 denote non-recurring pairs of points. $R$ is symmetric only if the chosen norm is symmetric. 

\begin{figure}[ht]
	\includegraphics[width=\columnwidth, scale=1] 
	{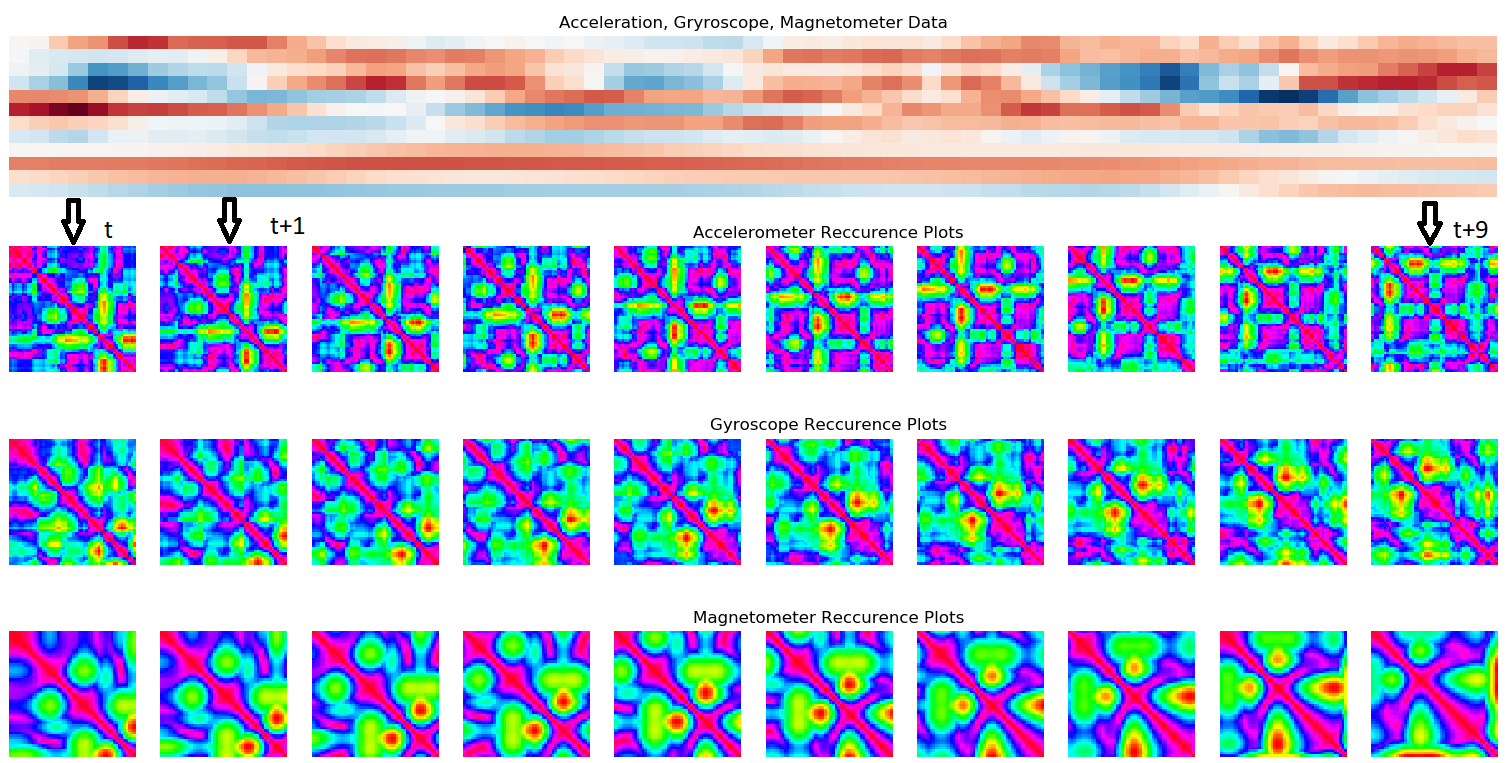}
	\caption{10 sequential recurrence plots from a accelerometer, gyroscope and magnetic sensor data streams.}
	\label{fig:rp}
\end{figure}

A {\it recurrence plot} (RP) is obtained by visualizing the recurrence
matrix~\cite{marwan2008}. 
Based on the simple estimation given by Eq.(\ref{eq:rp}), a powerful visual representation can capture the difference in dynamical behaviour. Figure~\ref{fig:rp} shows 10 sequential recurrence plots for a sliding window on $d$-dimensional time series, where $d$=12 is the dimension of data stream composed with accelerometer, gyroscope and magnetometer data.

\begin{figure*}[ht]
    \includegraphics[width = 2\columnwidth, scale=1] {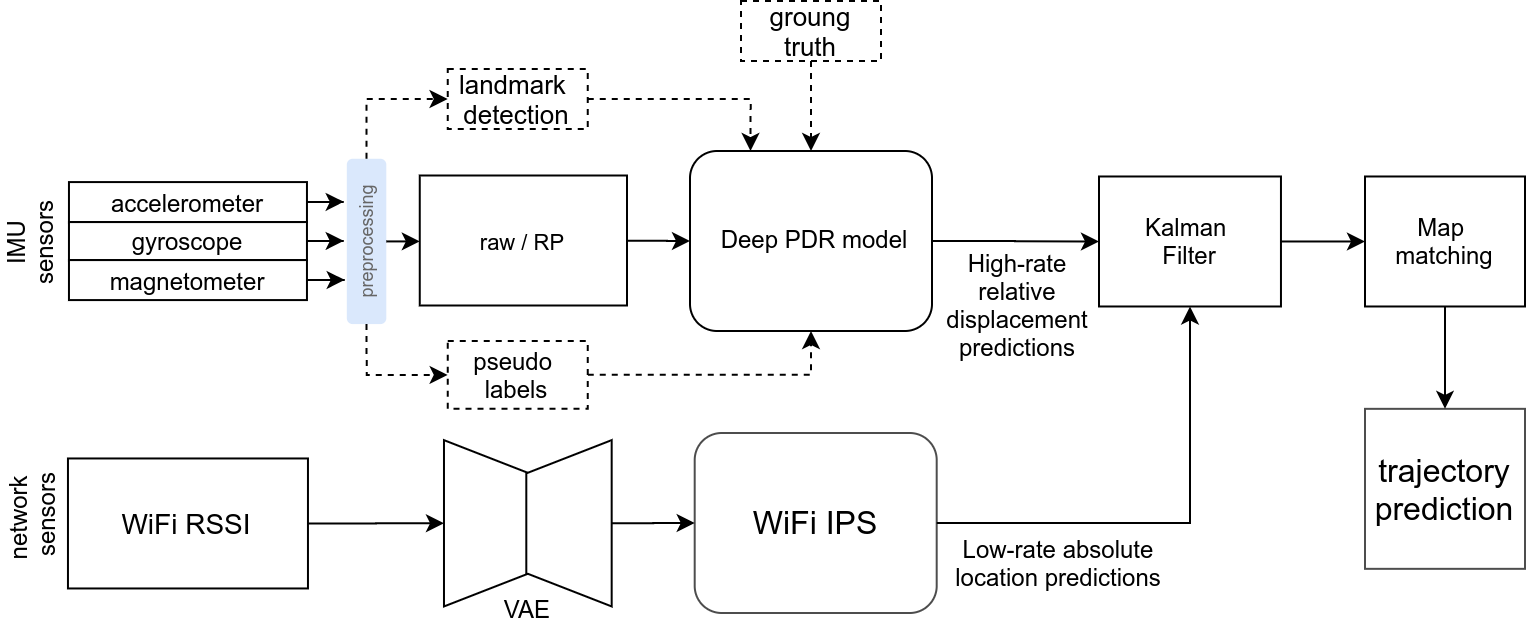}
    \caption{General overview of the architecture. Dashed boxes and lines represent data used to train the Deep PDR model.}
    \label{fig:architecture}
\end{figure*}

\subsection{System architecture}
\label{ssec:architecture}
Our system for indoor positioning and tracking is composed of four main components:
\begin{itemize}
	\item A deep PDR model that provides a high-rate update of the user's relative displacement.
	\item A WiFi fingerprinting component that provides a prediction of the absolute user's position each time a WiFi scan is received, which occurs approximately every 4 seconds.
	\item A Kalman filter to fuse the different rate predictions from the deep PDR and WiFi components. The filter provides an estimate of the user's position, without taking into account physical restrictions imposed by the environment.
	\item A map-free projection algorithm that projects the prediction from the Kalman filter on the paths that are possible given the physical constraints of the environment (corridors, doors, etc.). In this way, the final prediction is adjusted to a feasible route in which crossing regions of the environment that are impossible for a user on foot is avoided.
\end{itemize}

Figure~\ref{fig:architecture} shows the proposed architecture; the following sections describe the main components in detail.

\subsection{Deep PDR}
\label{ssec:deep_pdr}

Any PDR based system 
monitors the user's behavior by gathering relevant data from IMU sensors.
The sensors' data stream is processed into handcrafted features. Relevant and discriminative characteristics, such as number of steps, step length, orientation, etc, are extracted from the raw data.
	Finally, the user's relative position is using a theoretical dynamic model of the movement.

Classical PDR techniques need to infer the speed of the user through the determination of steps, extracted typically from accelerometer data, and an approximation of the user's step length. The error of PDR estimations is usually caused from both heading and step length error. The stride of the user does not have to be constant and depends, among other factors, on the physical characteristics of the user. 

Instead, we propose a deep learning approach to PDR. To learn the local displacement model from IMU sensor data, we make a simplifying assumption motivated by the analysis of the IPIN'19 challenge dataset. Indeed,
user tracks in the dataset
correspond to routes carried out inside administrative buildings composed mostly of long corridors. The landmarks provided with the data refer to user's direction changes. Therefore, we assume that there are no changes in orientation during the user's path between two consecutive landmarks, so the user moves in a straight line between these points. However, the user's speed is unknown and can vary due to an obstacle, such as a door or other people, or as a consequence of a user decision. We determine the speed by obtaining the number of steps from the data coming from the accelerometer, and adjusting the speed based on the distance between each two consecutive landmarks and their corresponding timestamps. In this way, the speed is not considered constant between two landmarks, but varies depending on the data provided by the accelerometer. 


\paragraph{CNN as Deep PDR model}
Our CNN consists of 3 convolution layers and 2 max-pooling layers followed by fully connected layers. Two dropouts layers interleaves convolution layers to improve the overall accuracy of CNN results.

CNN inputs the image-framed raw sensor data or RPs, and passes it through convolution layers. 
Convolution kernels in these layers vary in the function of input image size.
The pooling layer is placed after the activation of convolution layer. This layer can extract features of the convolution layer output by reducing the number of rows and columns of the image-like input. In our implementation, max pooling layer with a two by two filter (stride two) will store the maximum value of the two by two subsection. At the final stage of CNN, there are fully connected layers with softmax function which calculates the output of CNN. The softmax acts as a regressor based on the $(\Delta x, \Delta y)$ displacements. 

\paragraph{RNN as Deep PRD model}
We explore the performance of a deep PDR model by replacing CNN by a Recurrent Neural Network~\cite{rumelhart1986} (RNN). RNNs are specialized in processing sequences of values and capturing long-distance inter-dependencies in the input stream. They can pass information among time steps, which allows them to remember information about previous values in the sequence. When dealing with time series of IMU sensor data, recurrent networks are capable of identifying these temporal patterns and produce accurate predictions. In each step, the internal state of the RNN, a sort of 'memory' of previous time steps, is combined with the current time step input to produce an output. This way, the last output for a given sequence will be based on information obtained from all previous values in the sequence.

Vanilla RNN architecture suffers from some severe issues, like the vanishing and exploding gradient problems~\cite{bengio1994}, which makes optimization a complex challenge. Long Short Term Memory (LSTM) networks~\cite{hochreiter1997} have been designed as a way to avoid these problems while efficiently learning long-range dependencies. If fed in a bidirectional fashion, using both the data from start to end and from end to start, LSTMs can achieve better results, since they can recognize patterns in both directions. In our experiments, we use bidirectional LSTMs and assess their capacity to learn the user's relative displacement model given a series of raw sensor data. 



\subsection{Landmarks and pseudo labels}
\label{ssec:landmarks}
Landmarks play an important role in indoor positioning and tracking; they refer to direction changes, restrained passages like doors and elevators. Landmarks can be often identified by analysing sensor data. Once identified, they allow to obtain pseudo labels and thus generate a 
richer training set, which is critical for training an accurate deep PDR model.


Indeed, 
genuine ground truth annotations are sparse and mostly available for a limited number of landmarks. On the other hand, raw sensor data are massively generated at high rate. So we develop a method to annotate sensor data with pseudo labels and generate a large annotated set for training a deep PRD model. It is based on simpler tasks of user activity and landmark detection and a guess about how users behave between the landmarks. To generate the pseudo labels, we make a simplifying assumption that user moves along a straight line between any two landmarks. 

Such an assumption is validated in a major part of indoor environments where any user's trajectory can be represented by a sequence of segments and the error is limited to choices in multi-door passages, the width of corridor, etc. Once landmarks are identified, pseudo labels are obtained by interpolating user's position between two landmarks, under the assumption that all paths between landmarks are straight trajectories with no turns. 


We run a sliding window over the IMU data stream obtained from accelerometer, gyroscope and magnetometer data and associate every image-framed input with the corresponding change in user's position. Temporal and multi-modal correlations present in sensor data are learned using a deep PDR model. We train a network to predict the relative displacement using image-shaped input with associated ground truth or pseudo labels from the training set.

Using this approach, the inertial sensor readings are used to predict relative user's displacements. 
The challenge logfiles provide the orientation of the device. Since the user trajectories have been recorded while holding the phone in front of the user's chest, the provided yaw angle corresponds with the user's heading. All the data contained in the training and validation sets are used to train the deep learning model that will be responsible for predicting the user's trajectory based on the inertial sensor data, thus replacing the classic PDR method.

\begin{figure}[ht]
\includegraphics[width = 1.0\columnwidth, scale=1] {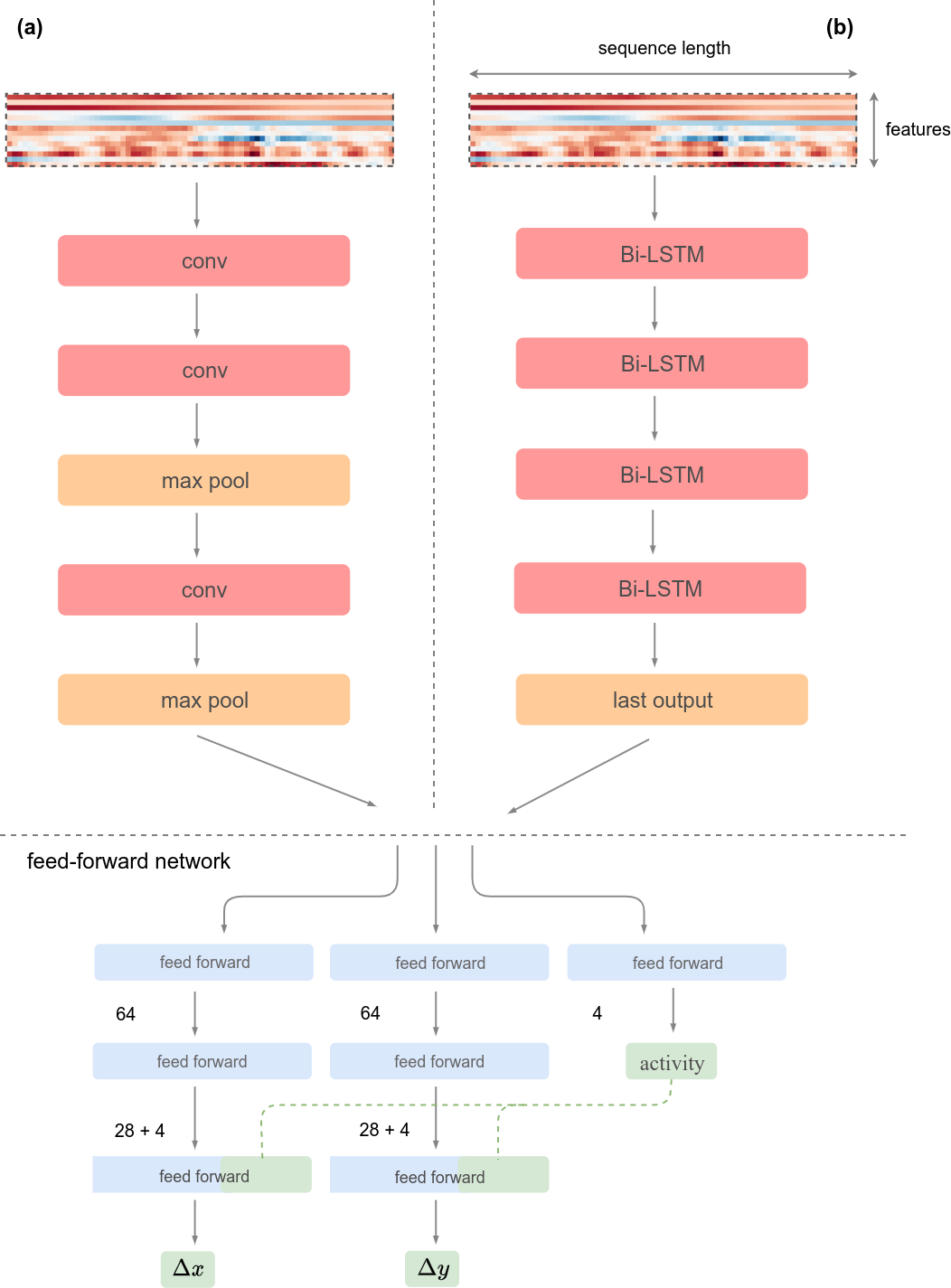}
\caption{Deep PDR network. (a) CNN based (b) RNN based.}
\end{figure}

\subsection{WiFi: VAE based predictions}
\label{ssec:wifi}
The deep PDR model predicts user's relative displacements and is prone to drift accumulation. To reduce the drift, we add the absolute position estimation to the system. We use available low frequency WiFi RSS data to build a WiFi based positioning~\cite{Ma2019wifi}.

%

While it is relatively easy to collect unlabeled WiFi data by crowdsourcing, it is significantly more expensive and tedious to annotate the data with an exact location. WiFi data is massively collected (every 4 seconds), but a small part is annotated with coordinates.  

The {\it semi-supervised learning} is a paradigm where both labeled and unlabeled data are used for building accurate prediction models. 
The semi-supervised setting is well suited for WiFi data collection where one or more equipped devices can combine a low cost collection of non-annotated WiFi data with a limited annotation effort. Several semi-supervised methods~\cite{chidlovskii19,ghourchian2017,pulkkinen11,yoo17} showed their efficiency in reducing the annotation needed for an accurate WiFi based localization. 

We follow~\cite{chidlovskii19} in applying the recent advances in deep and semi-supervised learning to the WiFi based positioning.
We implement a method based on Variational Auto-Encoder (VAE)~\cite{kingma2019} introduced in Section 2.1, that significantly reduces the need for the labeled data. It can combine a small amount of the labeled data with a large unlabeled dataset to build an accurate predictor for the localization component. We adapt the standard VAE encoder-decoder architecture, where the encoder maps the RSS data into latent variables, and also plays the additional role of a regressor of the available labeled data. The VAE decoder plays the regularization role on both labeled ans unlabeled data. 

The method is semi-supervised and able to train a prediction model from a small set of annotated WiFi observations (10-15\% of WiFi data used by the VAE) completed with a massive set of non-annotated WiFi observations. 

Relative user's displacements are predicted by the deep PRD model at high frequency, while Wifi-based absolute predictions are low frequent; the two predictions are fused by Kalman Filter. 

\subsection{Kalman Filter for fusion}
\label{ssec:kf}


Existing data fusion frameworks mainly include particle filter and Kalman filter~\cite{chen-2015-fusion,davidson-2017-survey,poulose-2019-sensor}. The particle filter may achieve reasonable accuracy by deploying a large number of particles, but a large amount of computational cost is required and is not suitable for resource limited smartphones.

The Kalman filter-based approaches are computational lightweight~\cite{hosseinyalamdary_deep_2018}. However, an explicit measurement equation connecting user's position with RSS measurements is unavailable due to complex indoor radio propagation, thus rendering the measurement noise statistics unavailable. Previous Kalman filter-based fusion approaches manually and empirically set the related measurement noise covariance matrix. As a result, the fusion process cannot adapt the uncertainty of WiFi positioning results and, thus, rendering a degraded positioning accuracy.

We follow~\cite{chen-2015-fusion} in adopting Kalman Filter as a sensor fusion framework for combining low-rate WiFi and high-rate PDR predictions. 
The sensor fusion problem is formulated in a linear perspective, so enabling the whole system to run on a smartphone.

\subsection{Map-free projection}
\label{ssec:project}
The Kalman Filter does not take into account physical constraints imposed by the floor layout. At the same time, the user may not go through walls, she may change the floor only at stairs, etc. Therefore, we performed an additional step of adjusting the Kalman Filter output by projecting its output on the walkable paths only. Even though that the floor map was not provided explicitly, we could implicitly reconstruct the underlying map by extracting 'walkable paths' between landmarks. Thus, as the final step, we could adjust the output of the Kalman filter based on the floor layout, by projecting its prediction to the closest path.

We introduce an additional component which turns to be critical in the regression based localization. All conventional regression methods often ignore the structure of the output variables and therefore face the problem of predictions outside the target space. Indeed, when testing our system on IPIN'19 dataset, a number of predictions fail to fit the indoor building space. We therefore implement a method of a structured regression~\cite{yoo17} which guarantees that predictions fit the feasibility space.  

A naive solution assumes an access to an accurate location map; then any location prediction is first tested for being inside the feasibility space, and a correction is required if the test fails. 
To make our system more generic and map-independent, we do not assume any map and count only on the training set for the possible corrections.

The method which turns to be robust in the semi-supervised setting, is based on the {\it weighted neighbourhood projection}. For each location prediction, we consider top $N_r$ neighbors in the available annotated set. The projection is given by the weighted sum of the neighbors; these weights are calculated as an inverse of distances between the prediction and corresponding neighbours. This projection belongs to a convex hull defined by the $N_r$ neighbours. 

The map-free projection works well when all neighbours are topologically close and the convex hull they define is a part of the feasibility space. However, if the neighbours are topologically distant (for example, located in different buildings), the error caused by the projection can increase. To minimize the risk of error, we consider rather small values of $N_r$. 



%% file: evaluation.tex
\section{Evaluation} 
\label{sec:eval}

\subsection{IPIN localization challenges}
\label{ssec:ipin}
The Indoor Positioning and Indoor Navigation (IPIN) conference
holds an annual competition that provides a rigorous evaluation methodology in order to fairly compare different technologies both in online (real-time, on-site) and offline (post-processing, off-site) settings~\cite{ipin_competition_2018}. 

\begin{figure}[ht]
    \includegraphics[width = \columnwidth, scale=1] {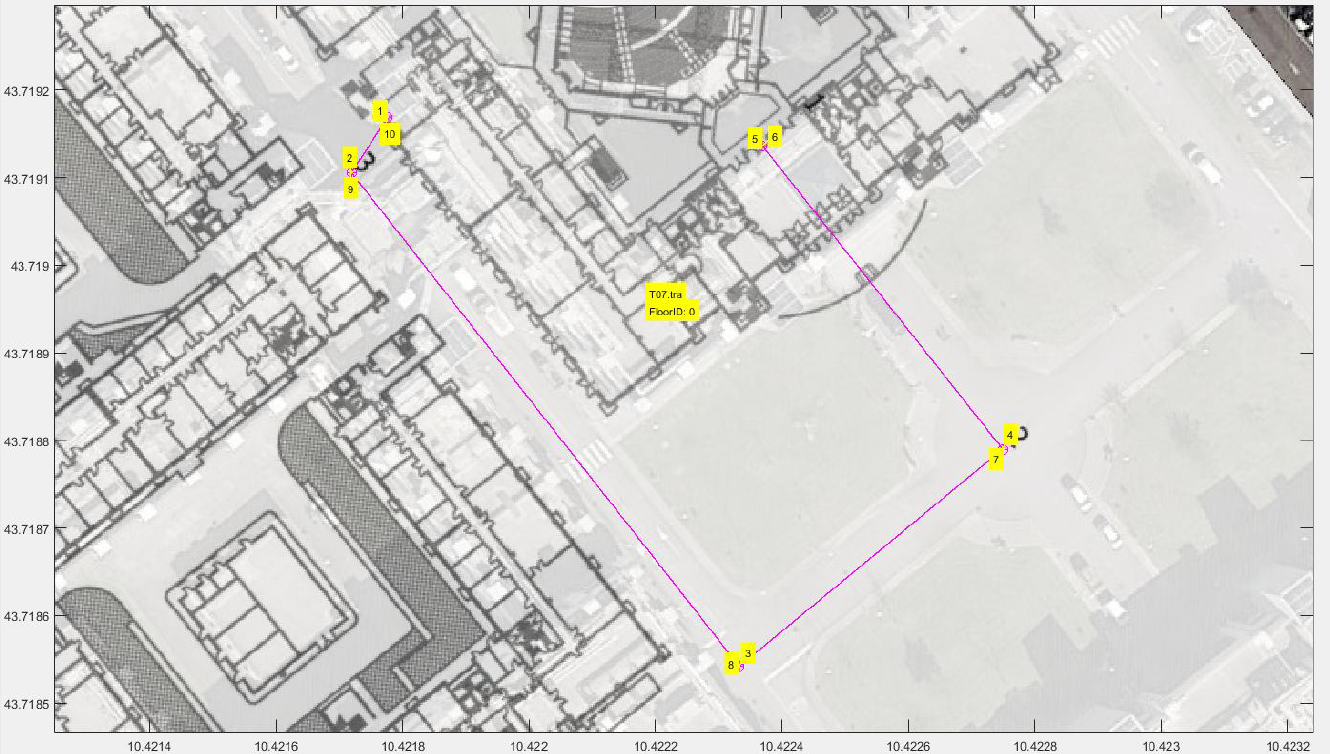}
    \caption{Example of a training path with annotated landmarks.}
    \label{fig:trainingpath}
\end{figure}

\begin{figure}[ht]
    \includegraphics[width = \columnwidth, scale=1] {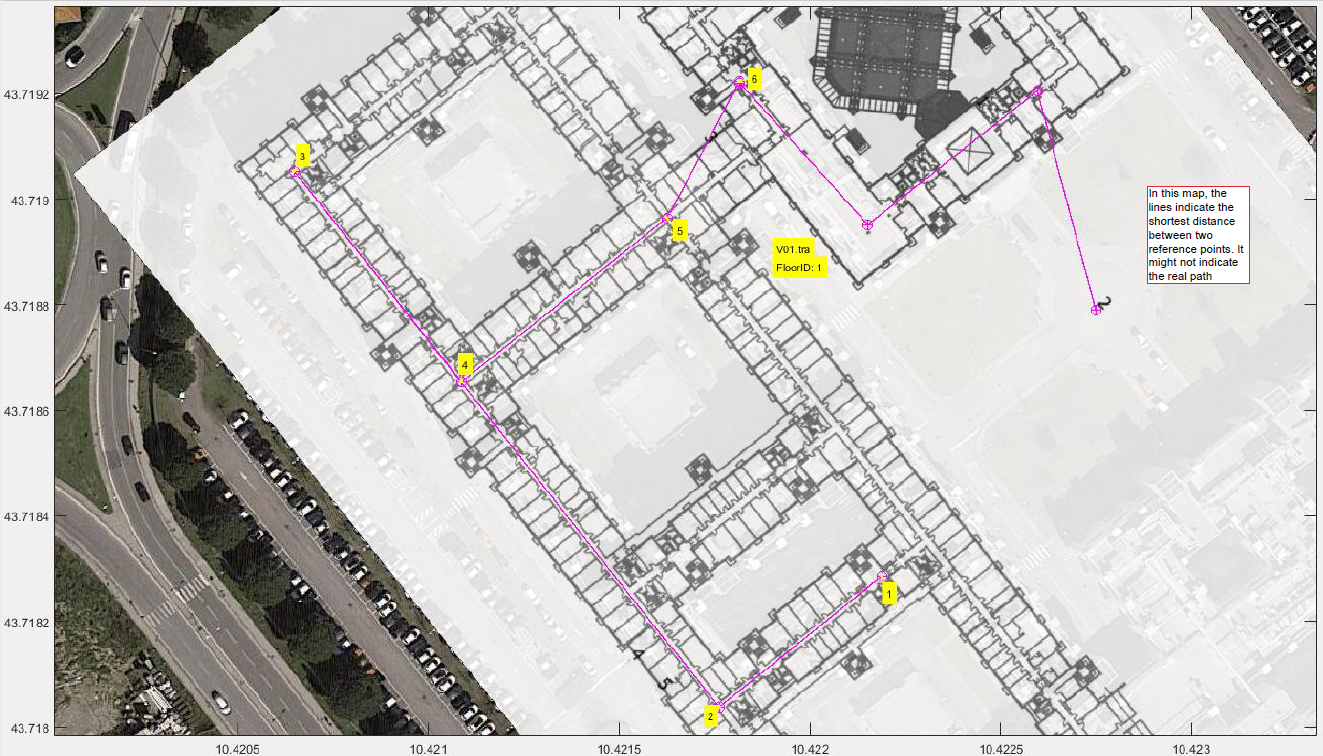}
    \caption{Example of a validation path with partially annotated landmarks.}
    \label{fig:validationpath}
\end{figure}

The main goal of the off-site smartphone based positioning track of the IPIN competition, is to recreate a path traversed by a person holding a conventional modern smartphone (Samsung A5, 2017), based on the readings from the smartphone's sensors. Sensors data was recorded and stored in a logfile using the "GetSensorData" Android application~\cite{android_app}. The application records all the raw data that is available from the smartphone sensors, such as WiFi/BLE RSS, GPS location, acceleration, gyroscope, magnetic field, orientation, pressure, light and sound intensity, etc.
%
The logfiles are divided into training, validation and evaluation sets. The organizers supplied a set of landmarks, consisting on user's positions at a given timestamp, for training and validation logfiles.
Training set consists of 50 logfiles corresponding to 15 different trajectories, of length \char`\~5 mins each, that were traversed multiple times in both directions (see Figure~\ref{fig:trainingpath}). Validation set contained 10 logfiles associated with 10 different trajectories, of lengths \char`\~10 mins each (see Figure~\ref{fig:validationpath}).
The main difference between training and validation logfiles is that in the training logfile, all significant turns have been recorded (i.e. annotated) with a landmark, while in the validation set a trajectory between two consecutive landmarks is not necessarily a straight line and may include turns, u-turns, stops and other challenging movements.

The evaluation logfile contains only recordings of the sensors data, for ~20 mins without any landmarks information. The goal of the competition is to recreate the path of the user, based on this sensors data, providing user position estimations every 0.5 seconds. 
The final results are benchmarked by the organizers, on landmarks unknown to the competitors and 75\% quartile of the error distribution is used to determine the winner. The competition data is publicly available~\cite{ipin_competition_2019}.


\subsection{Evaluation results} 
\label{ssec:results}

\begin{table}[th]
\centering
\begin{tabular}{c|c|c||c||c||l|l|l|l}
\multicolumn{5}{c||}{IPIN'19 Indoor}  & MAE & 50\%& 75\% & 90\% \\
\multicolumn{5}{c||}{Localization Challenge}&(m.) & Err & Err* & Err  \\  \hline \hline
\multicolumn{5}{c||}{Winner}   &  2.0 &  1.5  & {\bf 2.27} &  5.1 \\ \hline
\multicolumn{5}{c||}{2nd place(*)}&{\bf 1.7}&  {\bf 1.3}  & 2.36 & {\bf 3.9} \\ \hline
\multicolumn{5}{c||}{3rd place}&  2.1 &  1.8  & 2.54 &  \\ \hline \hline
\multicolumn{5}{c||}{Our pipeline}&\multicolumn{4}{c}{}  \\ \hline 
PLs & RPs &Model&Wi- &PRJ  & MAE  & 50\%& 75\% & 90\% \\
    &     &     &Fi  &     & (m.) & Err & Err* & Err   \\  \hline \hline
    &     & RNN &\chk&\chk & 1.79 & 1.33 & 2.44 & 4.50 \\ \hline
\checkmark&     & RNN &\checkmark&\checkmark& {\bf 1.53} & 1.29 & 1.92 & 3.31 \\ \hline
\chk&\chk & RNN &    &\chk& 2.10  & 1.56  & 2.85 & 4.49 \\ \hline
\chk&\chk & RNN &\chk&    & 1.74 & 1.47 & 2.19 & 3.32 \\ \hline
\chk&\chk & RNN &\chk&\chk& 1.64 & 1.28 & 1.99 & 3.45 \\ \hline 
\hline
    &     & CNN &\chk&\chk& 1.98 & 1.42 & 2.46 & 4.51 \\ \hline
\chk&     & CNN &\chk&\chk& 1.54 & 1.16 & 1.99 & {\bf 3.21} \\ \hline
\chk&\chk & CNN &    &\chk& 1.97 & 1.38 & 2.32 & 5.01 \\ \hline
\chk&\chk & CNN &\chk&    & 2.22 & 1.89 & 2.83 & 4.11 \\ \hline
\chk&\chk & CNN &\chk&\chk& 1.58 &{\bf 1.05}&{\bf 1.80} & 3.70 \\ \hline
\end{tabular}
\caption{The best results of IPIN'19 challenge and MAE, 50\%, 75\% and 90\% errors for our system, by ablating pseudo labels(PLs), RPs, WiFi and map-free projections(PRJ).}
\label{tab:error}
\end{table}

We validate the effectiveness of our system by ablating different components,  measuring the corresponding localization errors and comparing them to the challenge's best results. In all experiments,  
we evaluate 75\% quartile of the error distribution used by the IPIN'19 challenge organizers during the competition. 
In addition, we report the standard Mean Average Error (MAE), and 50\% and 90\% quartiles.

Table~\ref{tab:error} first reports three top results of the challenge. Then it presents results when using CNN and RNN as deep PDR models. We feed the network with raw sensor data streams or recurrence plots and ablate pseudo labels, WiFi and map-free projections.

The best performance of 1.80 m of 75\% error is obtained for CNN as deep PDR model, completed with pseudo labels and RPs. In IPIN'19 challenge, the winner reported 2.27 m error (our contribution with 2.36 m error took the 2nd place). In other words, our improved architecture allows to reduce the winner's error by 20\%. This improvement is obtained due to adding RPs, fine-tuning the full pipeline and hyper-parameter optimization.

Beyond the deep PDR, we also ablate WiFi and map-free projection components of our pipeline. As Table~\ref{tab:error} shows, both components play an important role. Localization and tracking without WiFi predictions or map-free projection lead to an important performance drop, for both CNN and RNN models. 
\begin{figure}[ht]
\includegraphics[width=\columnwidth,scale=1.0]{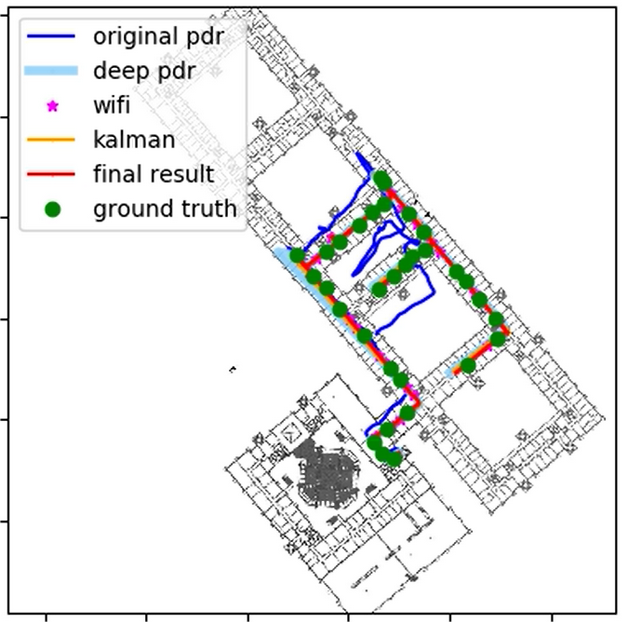}
\caption{Snapshot of the video comparing different scenarios.}
\label{fig:demo}
\end{figure}
\paragraph{Visual comparison.}
Beyond the evaluation results, we generated a video to visually compare the behaviour of different configurations of our network on the test tracks (this video is provided in Additional material). Figure~\ref{fig:demo} offers a snapshot of the video. For the test track, it shows standard and deep PDRs, WiFi based predictions, KF fusion of PDR and WiFi predictions, the final predictions after map-free projection, as well as the ground truth. 

The most remarkable is the difference between standard and deep PDRs, with the later showing a much smaller accumulated drift. Then, WiFi global position predictions and Kalman Filter fusion allow to correct errors of the deep PDR. Finally, the map-free projections allow to fix some impossible predictions and project them back into the feasible navigation space.

\subsection{Discussion and Future Work}
\label{ssec:discussion}

Most important lessons learnt from the evaluation are the following: 
\begin{enumerate}
\item Deep PDR represents a strong alternative to the standard PDR. Deep PDR models outperform them in all configurations and allow to reduce the accumulated drift. 
\item Using recurrence plots is preferable to a direct, naive conversion of sensor data in 2D image-shaped representation.
\item Magnetometer data turns to be valuable information for indoor positioning and tracking; in combination with accelerometer and gyroscope data, it contributes to the reducing localization error. Instead, removing magnetic field data leads to a performance drop.
\item Our attempt to take a benefit from sequential nature of sensor data and to deploy more complex LSTM as a deep PDR model was only partially successful. While the results outperform the last year competition winner, they were slightly worse than those we obtained with CNN, despite an intensive hyper-parameter optimization. 
It would be interesting to apply latest state-of-the-art techniques, such as attention mechanism to take advantage of the sequential nature of the sensors data. 
\end{enumerate}

Earlier we mentioned several contributions enabling our system to obtain the state of the art performance on IPIN'19 dataset.
Another important factor is an assumption about a corridor-based navigation space; it allows to simplify the landmark detection and the generation of pseudo labels and, therefore to train accurate deep PDR models. Instead, indoor positioning and tracking in multiple open spaces with erratic user navigation represents a more serious challenge. 

Relaxing the simplifying assumption represents the most intriguing direction of future work. 
One promising direction may come from our WiFi component which avoids pseudo labels; instead it deploys the semi-supervised learning to successfully project both labeled and unlabeled RSS WiFi data in VAE latent space and to make absolute position predictions.